\documentclass[10pt,twocolumn,twoside]{IEEEtran}

\usepackage{amsmath}
\usepackage{amssymb}
\usepackage{graphicx}
\usepackage{hyperref}
\usepackage{booktabs}
\usepackage{algorithm}
\usepackage{algorithmic}
\usepackage{listings}
\usepackage{xcolor}
\usepackage{tabularx}
\usepackage[utf8]{inputenc}
\usepackage{subcaption}
\usepackage{cite} 
\usepackage{ragged2e}

\begin{document}

\title{Optimizing MoE Routers: Design, Implementation, and Evaluation in Transformer Models}
\author{
  Dan Harvey (dyh2111) \\
  \and
  George Weale (gmw2143) \\
  \and
  Berk Yilmaz (by2385) \\
  \and
  DBGW \\
}
\maketitle

\begin{abstract}
Mixture of Experts (MoE) architectures increase large language model scalability, yet their performance depends on the router module that moves tokens to specialized experts. Bad routing can load imbalance and reduced accuracy. This project designed and implemented different router architectures within Transformer models to fix these limitations. We experimented with six distinct router variants Linear, Attention, Multi-Layer Perceptron (MLP), Hybrid, Hash, and our new MLP-Hadamard. We characterized these routers using BERT and the Qwen1.5-MoE model, looking at parameter efficiency, inference latency, routing entropy, and expert utilization patterns. Our evaluations showed distinct trade-offs: Linear routers offer speed, while MLP and Attention routers provide greater expressiveness. The MLP-Hadamard router shows a unique capability for structured, sparse routing. We successfully replaced and fine-tuned custom routers within the complex, quantized Qwen1.5-MoE model. This work provides a comparative analysis of MoE router designs and offers insights into optimizing their performance for efficient and effective large-scale model deployment.
\end{abstract}

\section{Introduction}

\subsection{Background}
Mixture of Experts (MoE) architectures scale large language models without proportional increases in computational costs during inference \cite{shazeer2017outrageously, fedus2022switch}. MoE models conditionally activate a parameter subset for each input token for each block. This increases parameter count while maintaining computational demands. A router component is the most important part of these architectures. The router determines expert (sub-network) processing for each token. Router component design impacts MoE model efficiency and effectiveness, as the router operates at each layer. This routing mechanism is a learned component within the neural architecture. It performs conditional computation to direct information flow through the network based on input features.

The Mixture-of-Experts (MoE) architecture has been the solution for scaling large language models (LLMs) efficiently. By routing tokens to specialized expert modules, MoEs allow models to maintain capacity while only activating a sparse subset of parameters during inference, reducing computational cost. However, the performance of MoEs is highly dependent on the design of the router module, which determines how tokens are dispatched to experts.

In this report we designed and evaluated router variants. These variants showed differences in complexity, determinism, and trainability. The report looked the following router variants: LinearRouter, a linear projection; AttentionRouter, a softmax-based attention mechanism; MLP Router, a multi-layer perceptron with non-linear activations; HybridRouter, a combination of linear and attention mechanisms; and HadamardRouter, deterministic routing using a Hadamard matrix, and parameter-free hash-based deterministic routing.

\subsection{Problem Statement \& Motivation}
MoE models give efficiency advantages by having access to many expert networks while only activating a few \cite{jiang2024mixtralexperts}. However, their practical performance is often limited by suboptimal routing mechanisms, this causes poor accuracy and increased inference latency. Most MoE implementations use a linear projection followed by a softmax to compute routing probabilities, which cannot capture complex relationships between token representations and expert specializations. Load imbalance is where certain experts are over utilized while others remain underutilized, this is a large challenge to achieve the full efficiency potential of MoE architectures. From a deep learning perspective, this is the main neural network design challenge: how to optimally split up the feature space so each expert can specialize effectively while maintaining balanced utilization. This project tries to fix these problems by looking at different router designs that can make more informed routing decisions while maintaining computational efficiency.

\subsection{Objectives \& Contributions}

The main goals of this project are: to design and build several router types for Mixture of Experts (MoE) models, using ideas from class; to make a flexible system for checking and comparing these router designs using good neural network study methods; to study the speed, efficiency, and how well work is spread by different router types using careful math methods; and to find the best router designs for different use cases by carefully checking how they represent information.

Our main work is the creation and testing of six different routers (linear, attention-based, MLP-based, hybrid, hash, and hadamard), each showing a different way for information to move in the network; a flexible MoE layer setup that can work with different router types, which lets us try out different network setups easily; a way to check router performance using measures for information and how work is spread out; and a close look at what different routers can represent and how much computing they need for different ways of routing information, especially for large language models.

\section{Related Work}
\label{sec:related_work}

Shazeer et al. \cite{shazeer2017outrageously} first showed a sparsely-gated MoE layer, where a gating network, typically a linear layer followed by a softmax, determines which experts process each part of the input. This architecture shows the viability of scaling neural networks to "outrageously large" sizes by activating only a subset of parameters per token.

Fedus et al. \cite{fedus2022switch} further developed this with Switch Transformers, scaling MoE models to trillion parameters. Their router, similar to Shazeer et al., employed a linear projection and softmax, but adding an auxiliary load balancing loss. This loss term helps create a more uniform expert utilization, as it helps the challenge in MoE training where some experts become over-utilized while others are neglected. The Switch Transformer also simplified the MoE design by routing each token to only a single expert (top-1 routing), different to earlier designs that might combine outputs from multiple experts.

Researchers have also looked at more interesting routing mechanisms. Roller et al. \cite{roller2021hash} looked at hash-based routing as a parameter-free alternative. Hash layers map inputs to experts using hash functions, this gives a deterministic and computationally inexpensive routing mechanism. However, the fixed nature of hashing will limit adaptability and load balancing without additional mechanisms.

Puigcerver et al. \cite{puigcerver2023poolingformer} explored "soft" mixtures of experts, moving away from hard, sparse gating towards mechanisms that allow for smoother transitions and combinations of expert contributions, potentially improving training stability and representational power. Their work, along with others, looked at the trade-offs between sparsity, expressiveness, and trainability in router design.

The Mixtral of Experts model \cite{jiang2024mixtralexperts} by Mistral AI represents a recent successful application of MoE principles in large language models. The router in Mixtral selects the top-k experts, a now common strategy to balance sparsity with the capacity to leverage multiple specialized networks.

Challenges in MoE training and routing, such as load imbalance, expert specialization, and the computational overhead of complex routers, remain active areas of research. This project builds upon these prior efforts by designing, implementing, and testing a set of router architectures. Our work differs by providing a direct comparative analysis of these varied router types within a standardized framework. We extend previous studies by trying out new combinations like the MLP-Hadamard router and by adding our analysis of the trade-offs inherent in different routing strategies.

\section{Methodology and Implementation Details}

\subsection{Overall System/Model Architecture}
At the highest level, the system has these components: (1) Router Implementations—a collection of different router architectures that inherit from a common base class and implement various neural routing mechanisms; (2) MoE Layer—a modular implementation of the Mixture of Experts layer that can use any router implementation, serving as the core conditional computation component; and (3) Evaluation Framework—tools for testing and comparing different router designs through principled deep learning metrics.

Figure 1 below shows the system architecture and the interaction between its components.

\begin{figure*}[h]
    \centering
    \includegraphics[width=1.8\columnwidth]{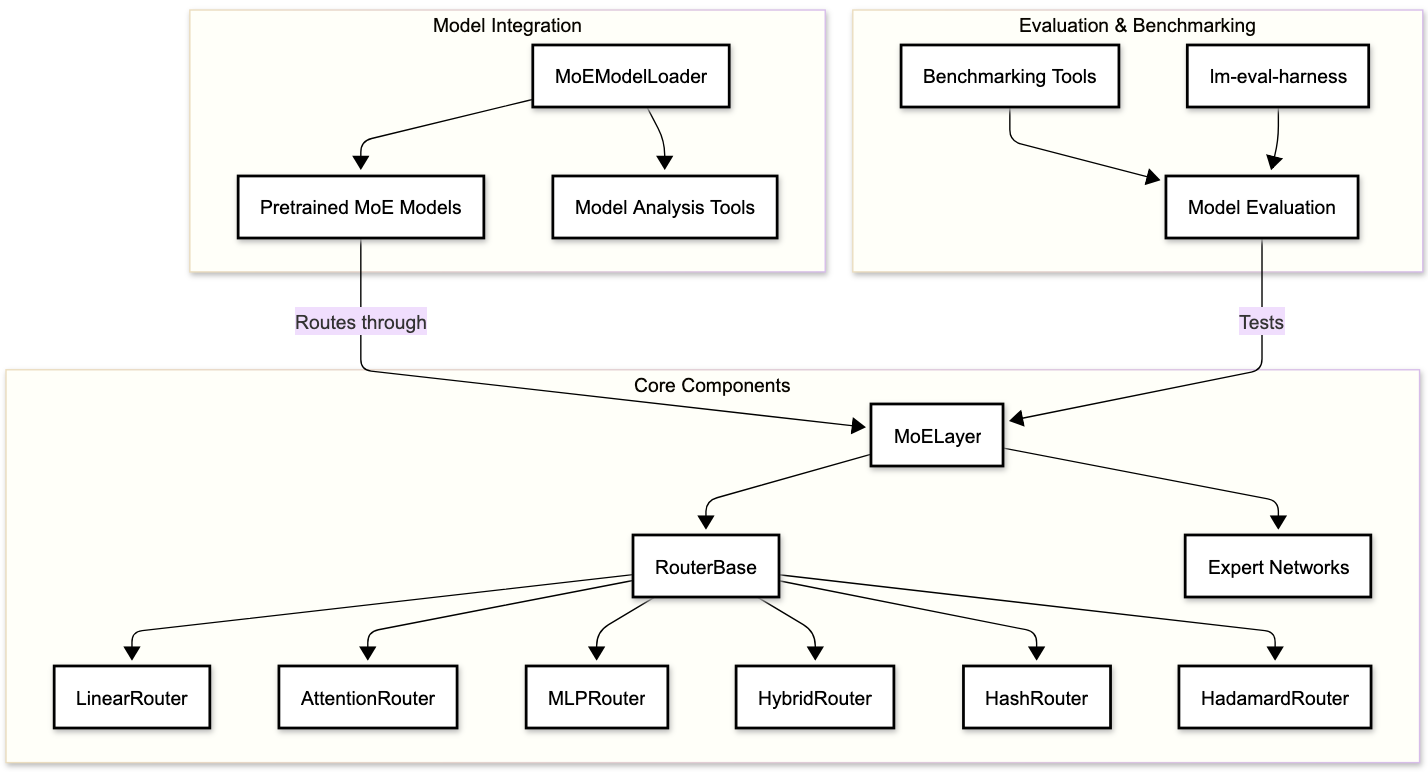}
    \caption{High-level architecture of the MoE system with modular router components}
    \label{fig:system_architecture}
\end{figure*}

\subsection{Key Components/Modules}

\subsubsection{Router Base Class}
We created a Router Base that that serves as the foundation for all router implementations. It provides a common interface and shared functionality. The core method that all router implementations must override is \texttt{compute\_router\_probabilities}, which takes token hidden states as input and returns routing probabilities for each expert.

\subsubsection{Linear Router}
We first started with a linear router. It is a single linear projection from the hidden dimension to the number of experts, followed by a softmax activation. This is:

\begin{equation}
p(e|x) = \text{softmax}(W \cdot x + b)
\end{equation}

where $x \in \mathbb{R}^d$ is the token representation, $W \in \mathbb{R}^{n \times d}$ is the weight matrix, $b \in \mathbb{R}^n$ is an optional bias term, and $p(e|x) \in \mathbb{R}^n$ is the probability distribution over $n$ experts. The linear projection shows the simplest form of feature extraction for routing, by computing a similarity score between the input and each expert's weight vector through an inner product. This is equivalent to a single-layer perceptron without activation function before the softmax, providing a baseline representational capacity.

\subsubsection{Attention Router}
The attention router adds transformer architectures—the attention mechanism—to the routing problem. It treats token embeddings as queries and learns expert embeddings as keys. This allows for more expressive routing decisions through the powerful inductive biases of attention. This is shown as:

\begin{equation}
\text{scores}(q, k) = \frac{q \cdot k^T}{\sqrt{d_k} \cdot \tau}
\end{equation}

\begin{equation}
p(e|x) = \text{softmax}(\text{scores}(q, k))
\end{equation}

where $q = W_q \cdot x \in \mathbb{R}^{d_k}$ represents token queries through a learned projection $W_q \in \mathbb{R}^{d_k \times d}$, $k \in \mathbb{R}^{n \times d_k}$ represents expert keys stored as learnable parameters, $d_k$ is the dimension of the key vectors, and $\tau$ is a temperature parameter that controls the sharpness of the distribution. Experts can be represented as distinct embeddings in a learned space, allowing them to develop specialized "profiles" that tokens can match against.

\subsubsection{MLP Router}
The MLP router extends the linear router by adding a hidden layer with non-linear activation, significantly increasing the expressivity of the routing function. The multi-layer perceptron architecture allows learning more complex routing patterns through non-linear feature transformations:

\begin{equation}
h = \sigma(W_1 \cdot x + b_1)
\end{equation}

\begin{equation}
p(e|x) = \text{softmax}(W_2 \cdot h + b_2)
\end{equation}

where $\sigma$ is a non-linear activation function, $h \in \mathbb{R}^{d_h}$ is the hidden representation, and $W_1 \in \mathbb{R}^{d_h \times d}$, $W_2 \in \mathbb{R}^{n \times d_h}$, $b_1 \in \mathbb{R}^{d_h}$, $b_2 \in \mathbb{R}^n$ are learnable parameters. The non-linearity allows the router to partition the feature space in ways that are not possible with linear projections alone, adding a more expressive routing decisions based on complex feature interactions.

\subsubsection{Hybrid Router}
The Hybrid Router combines linear and attention-based routing mechanisms. It computes routing scores through both methods and combines them using a learned or fixed weighting parameter:

\subsubsection{MLP Hadamard Router}
We were curious about looking at different feature interactions for routing beyond simple concatenation or summation as in hybrid approaches, thus we implemented an MLP Hadamard router. This router first processes the input token representation $x \in \mathbb{R}^d$ through a standard MLP layer to get an intermediate representation $h_{mlp} \in \mathbb{R}^{d_h}$. This is then combined with the original input token representation $x$ (or a projection of it) using an element-wise Hadamard product ($\odot$). The idea of this is that the MLP can learn to extract features that, when multiplied with the original features, highlight or suppress specific aspects relevant for routing.

\begin{equation}
h_{mlp} = \sigma(W_1 \cdot x + b_1)
\end{equation}

where $W_1 \in \mathbb{R}^{d_h \times d}$ and $b_1 \in \mathbb{R}^{d_h}$. If $d_h \neq d$, $x$ might be projected to $d_h$ using $W_p \in \mathbb{R}^{d_h \times d}$: $x_p = W_p \cdot x$. The Hadamard product is then:

\begin{equation}
h_{hadamard} = h_{mlp} \odot x_p \quad (\text{or } h_{mlp} \odot x \text{ if } d_h=d)
\end{equation}

The resulting vector $h_{hadamard} \in \mathbb{R}^{d_h}$ is passed through another linear layer to produce the expert logits:
\begin{equation}
p(e|x) = \text{softmax}(W_2 \cdot h_{hadamard} + b_2)
\end{equation}

where $W_2 \in \mathbb{R}^{n \times d_h}$ and $b_2 \in \mathbb{R}^{n}$. This goal is to create an expressive routing function than a standard MLP alone.

\subsubsection{Hash Router}
We implemented a hash router, as we have seen this to be extremely fast (although in our implementation was the slowest by an order of magnitude). This router uses a hashing function to deterministically map input tokens to experts. Unlike other routers that learn routing policies, the hash router relies on a fixed, non-learned assignment strategy. We implemented this by applying a chosen hash function to the token representation \(x\) (or a derivative of it) to produce a hash value. This value is then mapped to an expert index, using the modulo operator with the total number of experts \(E\):
\begin{equation}
\text{hash\_value} = \text{HashFunction}(x)
\end{equation}
\begin{equation}
\text{expert\_index} = \text{hash\_value} \pmod{E}
\end{equation}
The resulting routing "probability" distribution \(p(e|x)\) for a given token is effectively a one-hot vector, assigning a probability of 1 to the selected expert and 0 to all others:
\begin{equation}
p(e_i|x) = \begin{cases} 1 & \text{if } i = \text{expert\_index} \\ 0 & \text{otherwise} \end{cases}
\end{equation}
In theory, the advantage of this approach is its computational speed and simplicity, as it bypasses the need for learnable parameters and calculations in the routing step itself. Its performance is dependent on the hash function's ability to distribute tokens well, and does not have the adaptability that the learned routers have. Ensuring the load balance was challenging as the hash function easily adapts to the parameter distribution.

\subsubsection{MoE Layer}
The MoE layer is a modular layer for any of the router implementations. It handles routing tokens to experts based on the router's decisions, combining outputs from multiple experts using weighted aggregation, and computing load balancing loss to make sure uniform expert utilization. The MoE layer is designed to be a drop-in replacement for a standard feed-forward network in transformer models, keeping the same input and output dimensions while using a specialized internal structure that enables conditional computation through expert selection.

\subsubsection{Base MoE Model - BERT}
To characterize our constructed routers, we needed a reproducible and publicly available testbed that would allow us to apply each router in conjunction with a feed-forward network (FFN) in a transformer-like setting. For this, we used BERT (Bidirectional Encoder Representations from Transformers) as the foundation of our evaluation framework \cite{devlin2019bert}.

BERT’s architecture is relatively simple compared to more recent decoder-only models, making it easier to integrate custom router components. Its pre-training objectives—masked language modeling (MLM) and next sentence prediction (NSP)—allow BERT to learn token representations. BERT models are readily available as pre-trained checkpoints through the Hugging Face model hub, which gave us a starting point without the need for training.

By using BERT’s token embeddings as input to our router modules, we made sure that the routing evaluation was done under realistic activation patterns, rather than random tensors. We used these contextual embeddings to see how different routers handle token interactions.

We took out the feed forward network weights from the first encoder layer of a pre-trained bert-base-uncased model. We copied the parameters from the intermediate.dense and output.dense layers into the expert networks of our MoE modules, this weight initialization made sure that the MoE experts operated with the real learned parameters.

\subsubsection{Base MoE Model - Qwen}
We also chose \texttt{Qwen/Qwen1.5-MoE-A2.7B-Chat-GPTQ-Int4} as another base model to work with as well after seeing our sucess with BERT. Its MoE architecture is publicly available and characterized, providing a good testing are for router designs. The ``MoE-A2.7B'' means the model activates 2.7 billion parameters per token during inference, while its total parameter count is 14.3 billion. This architecture has 60 expert networks. The router selects the top \(k=4\) experts to process each token.

Qwen1.5 is a decoder-only Transformer model. Qwen1.5-MoE uses an MoE layer instead of standard Feed-Forward Networks (FFNs). This layer has the router and 60 FFN experts. Each expert is a SwiGLU (Swish Gated Linear Unit) FFN. It takes an input \(x_{expert}\) (output from the preceding layer, routed to this expert) and applies transformations defined by weight matrices \(W, V, W_2\). The formulation is \( \text{SwiGLU}(x_{expert}, W, V, W_2) = (\text{SiLU}(x_{expert}W) \odot x_{expert}V)W_2 \). This provides non-linear transformations. Qwen models use RMSNorm (Root Mean Square Layer Normalization) to stabilize training and improve performance. RMSNorm is applied before attention and FFN sub-layers. For an input vector \(x \in \mathbb{R}^{D_{model}}\), RMSNorm calculation is \( \text{RMSNorm}(x) = \frac{x}{\sqrt{\frac{1}{D_{model}}\sum_{i=1}^{D_{model}} x_i^2 + \epsilon}} \cdot g \). Here, \(g\) is a learnable gain parameter and \(\epsilon\) is a constant for numerical stability. Rotary Positional Embeddings (RoPE) add sequence order information. RoPE modifies query \(q_m\) and key \(k_n\) vectors at positions \(m\) and \(n\) by applying rotation matrices \(R_{\Theta, m}\) and \(R_{\Theta, n}\): \( q'_m = R_{\Theta, m} q_m \) and \( k'_n = R_{\Theta, n} k_n \). This encodes positional information within the self-attention mechanism. The Qwen1.5-MoE-A2.7B router is a linear projection \( W_{route} \in \mathbb{R}^{D_{model} \times N_e} \) (\(N_e=60\) experts) from the model hidden dimension to the expert count. This is followed by a softmax activation and top-4 selection, providing a baseline.

The ``-Chat'' part mean that we are using the model that has instruction-tuning and fine-tuning for conversational tasks. The ``-GPTQ-Int4'' part means that the model has had post-training quantization by the GPTQ algorithm to 4-bit integer precision. We did this so we would able to run this on a single A100 and even an L4.

\subsection{Data Flow}
\begin{figure}[h]
    \centering
    \includegraphics[width=0.9\columnwidth]{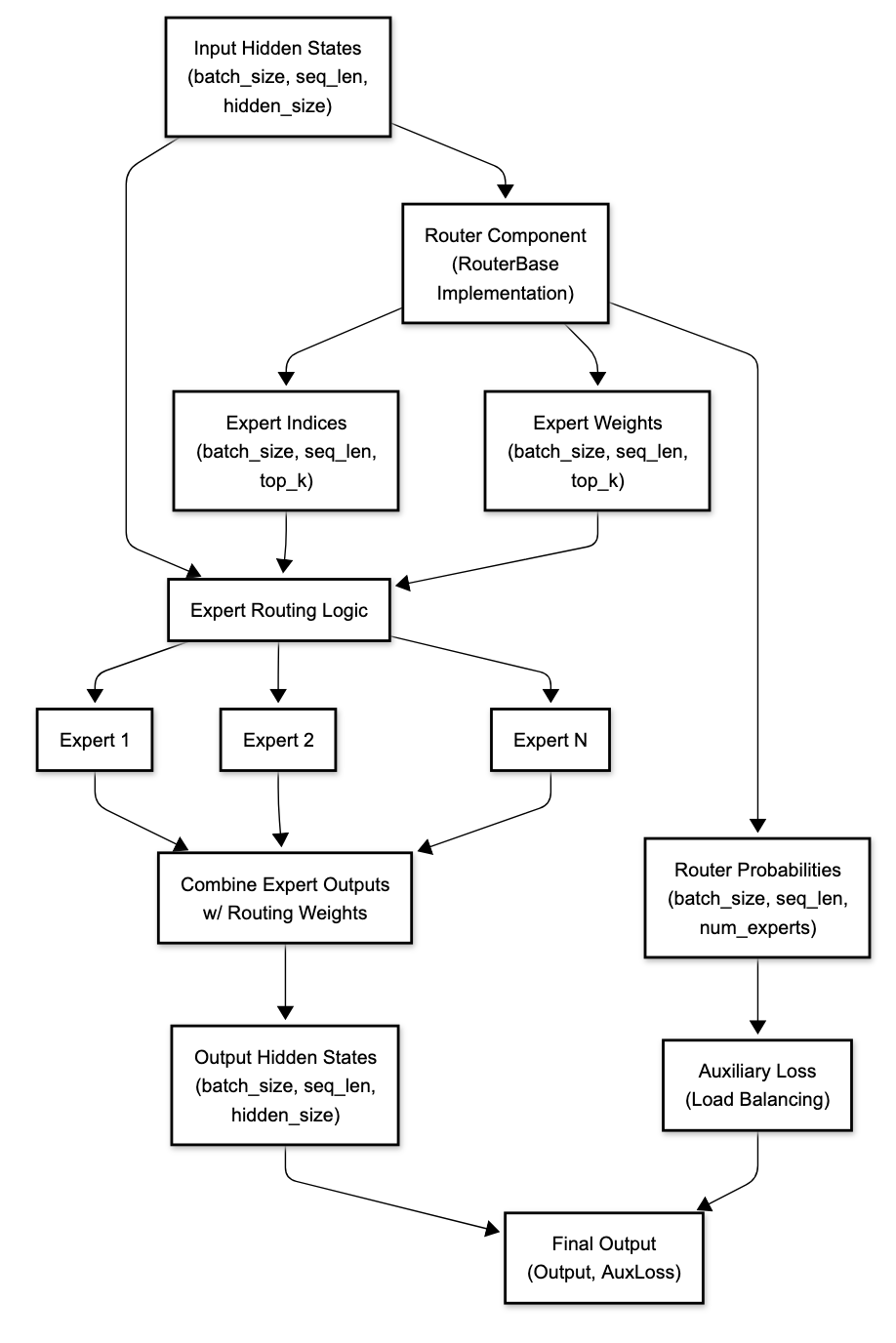}
    \caption{Data flow through core components in the forward pass}
    \label{fig:data_flow}
\end{figure}

The system is made to work with token representations from transformer models, typically with shapes $(batch\_size, sequence\_length, hidden\_size)$. The representation space in which routing occurs is the same high-dimensional embedding space used throughout transformer models, making this a natural extension of the standard architecture.

\subsection{Formulation}

\subsubsection{Router Selection Process}
Given token representations $X \in \mathbb{R}^{B \times S \times H}$ (where $B$ is batch size, $S$ is sequence length, and $H$ is hidden dimension), the router computes probabilities $P \in \mathbb{R}^{B \times S \times E}$ (where $E$ is the number of experts) and selects the top-$k$ experts for each token through a differentiable approximation of argmax:

\begin{equation}
\text{indices}_{i,j} = \text{topk}(P_{i,j}, k)
\end{equation}

\begin{equation}
\text{weights}_{i,j} = P_{i,j,\text{indices}_{i,j}}
\end{equation}

\begin{equation}
\text{weights}_{i,j} = \frac{\text{weights}_{i,j}}{\sum_{l=1}^{k} \text{weights}_{i,j,l}}
\end{equation}

where $\text{topk}(P_{i,j}, k)$ returns the indices of the $k$ highest values in $P_{i,j}$, and the weights are normalized to sum to 1. The top-$k$ selection adds a non-differentiable operation in the forward pass, but gradients can still flow through the selected experts using the straight-through estimator.

\subsubsection{Auxiliary Load Balancing Loss}

\begin{figure}[h]
    \centering
    \includegraphics[width=0.9\columnwidth]{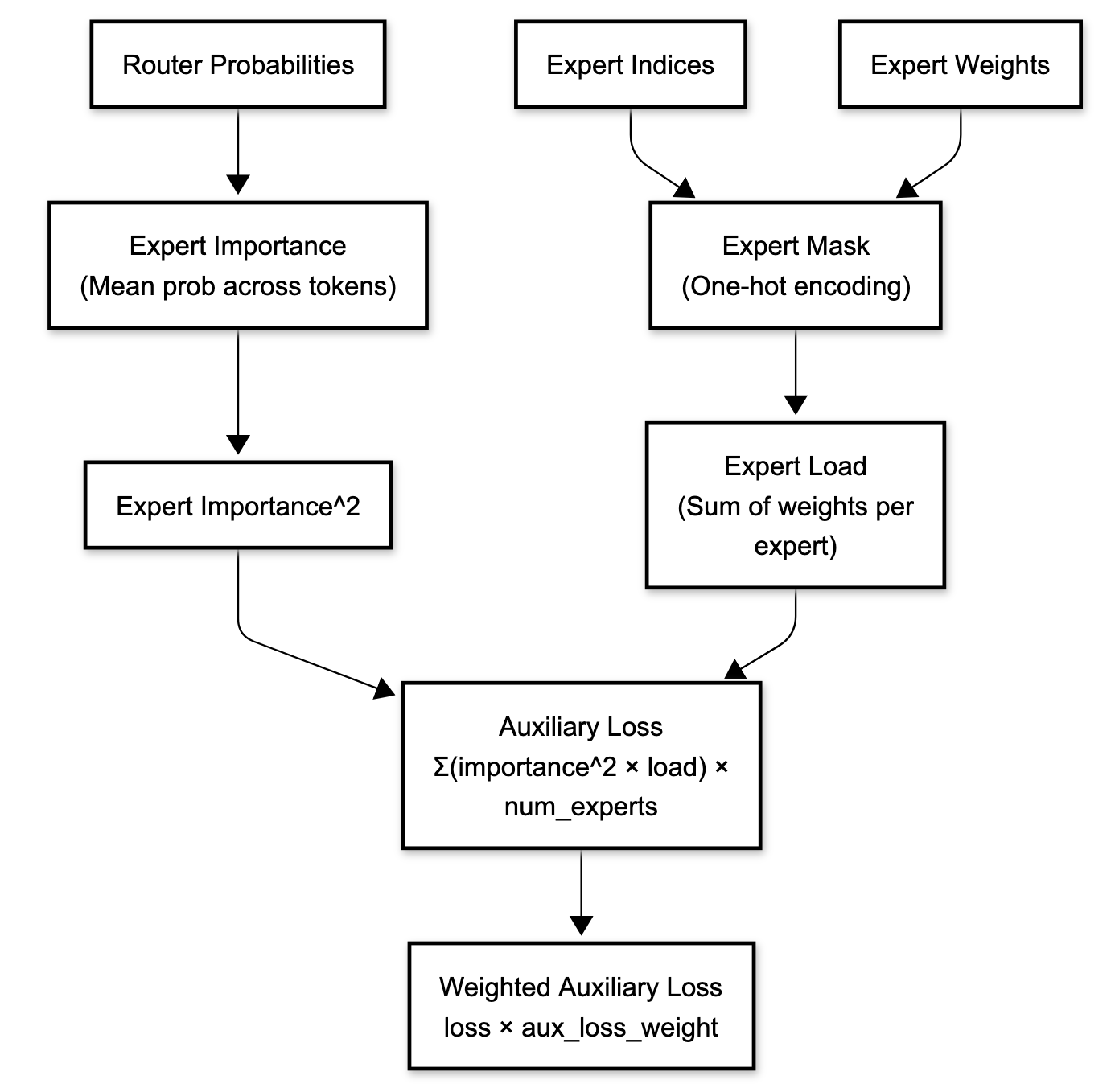}
    \caption{Auxiliary Loss Mechanism}
    \label{fig:aux_loss}
\end{figure}

To have balanced expert utilization, we calculate the auxiliary load balancing loss. Following \cite{fedus2022switch, shazeer2017outrageously}, this loss aims to distribute the computational load more evenly. First, we define the fraction of tokens dispatched to expert $e$ across a batch, 

$\text{expert\_load}_e$:
\begin{equation}
f_e = \frac{1}{B \cdot S} \sum_{i=1}^{B} \sum_{j=1}^{S} \sum_{l=1}^{k} \mathbb{I}((\text{indices}_{i,j})_l = e)
\end{equation}

where $\mathbb{I}(\cdot)$ is the indicator function. This $f_e$ corresponds to 
$\text{expert\_load}_e$ in your text if $\text{mask}_{i,j}$ is summed over experts.

The average routing probability (or importance) assigned to expert $e$:

\begin{equation}
g_e = \frac{1}{B \cdot S} \sum_{i=1}^{B} \sum_{j=1}^{S} P_{i,j,e}
\end{equation}

This $g_e$ corresponds to your $\text{expert\_importance}_e$. The auxiliary loss $\mathcal{L}_{\text{aux}}$ is then formulated as the scaled dot product of these two quantities, summed over experts:

\begin{equation}
\mathcal{L}_{\text{aux}} = \alpha_{aux} \cdot E \cdot \sum_{e=1}^{E} f_e \cdot g_e
\end{equation}

where $\alpha_{aux}$ is a hyperparameter. The original Switch Transformer paper \cite{fedus2022switch} uses a slightly different formulation that involves the square of the importance, which you have:

\begin{equation}
\mathcal{L}_{\text{aux}} = \alpha_{aux} \cdot E \cdot \sum_{e=1}^{E} (\text{expert\_importance}_e)^2 \cdot \text{expert\_load}_e
\end{equation}

This penalizes both imbalance in the number of tokens routed to each  expert (via $\text{expert\_load}_e$) and imbalance in the router's confidence for each expert (via $\text{expert\_importance}_e$). The quadratic term discourages the router from assigning high probabilities to only a small subset of experts.

\subsection{Experimental Setup / Training Process}

\subsubsection{Benchmark Setup}
All router evaluations were conducted using Google Colab Pro+, using a GPU-accelerated environment for performance measurements. We tried to use GCP, but were not able to get an A100 quota. The codebase was done in Python 3.11.12. Each router is in a modular class structure imported by cloning our GitHub repository into the working directory. The instance was run on Ubuntu 22.04 (Linux Kernel 6.1) as the operating system, a 6-core, 12-thread CPU, 84 GB of RAM, with an NVIDIA A100 40GB GPU with CUDA support enabled. 

\subsubsection{Base Models and Datasets}
We use two base models for experiments: BERT \cite{devlin2019bert} and Qwen1.5-MoE-A2.7B \cite{jiang2024mixtralexperts}. BERT gives a transformer architecture for router characterization with its learned token embeddings. The Qwen1.5-MoE model, uses a native MoE architecture, and is small and good for router performance evaluation and fine-tuning. We add the routers by replacing gating mechanisms in these models. For fine-tuning tasks during router replacement and adaptation functionality, we use the Tiny Shakespeare dataset.

\subsubsection{Evaluation Metrics}
Expert utilization analysis determines how routers distribute tokens across experts. This is visualizing the token distribution to identify biases or imbalances, calculating routing entropy as \(H(P) = -\sum_{e=1}^{E} p_e \log p_e\) where \(p_e\) is the proportion of tokens routed to expert \(e\), and looking at the auxiliary load balancing loss \(\mathcal{L}_{\text{aux}}\) as an sign of expert engagement. We measure token latency as the time for a single token to pass through the router and selected expert(s) during inference, benchmarked with a batch size of 1 to isolate per-token processing overhead. We also looked at router output characteristics, including mean top-k probabilities and variance of routing scores, to give insights into router behavior.

\subsubsection{BERT - Random Initialization Evaluation}
In the initial round of evaluations, we tested all routers using randomly initialized hidden states to establish a performance baseline. Under these  conditions, the routers showed expected behavior patterns, with lightweight designs like the LinearRouter achieving the fastest inference times, while more complex architectures such as the MLP and Hybrid routers demonstrated greater routing diversity. Although effective for verifying implementation correctness, random activations lacked the contextual richness needed to fully assess router behavior in realistic settings.

\subsection{BERT - Pretrained Initialization Evaluation}
To simulate real-world usage, we conducted a second round of evaluations using hidden states derived from a pretrained BERT model. With contextualized token embeddings, routers exhibited more distinct and polarized routing patterns, showing the meaningful structure of the input data. Output variance increased across all routers, with the MLP router showing the largest shift, highlighting its capacity to leverage pre-trained knowledge for nuanced expert selection. Expert utilization became more concentrated, showing how certain routers prioritize specific experts when handling semantically rich inputs. This BERT-based evaluation provided a more realistic and informative measure of each router's practical routing dynamics.

\subsection{Fine-Tuning the Linear Router}
We load a Qwen/Qwen1.5-MoE-A2.7B checkpoint in 16-bit precision, distributed over available hardware. We look for each Mixture-of-Experts layer by the \texttt{.experts} and \texttt{.router} attributes, look the expert's input size, and replaces the existing gating router with our \texttt{LinearRouter}. The model is placed into a Parameter-Efficient Fine-Tuning (PEFT) LoRA wrapper to avoid tuning all 14 billion parameters. We update only the four per-attention-block projection matrices (q\_proj, k\_proj, v\_proj, o\_proj) using low-rank (rank = 8) residual adapters with \(\alpha=32\) and 5 percent dropout. This LoRA method reduces trainable parameters to several hundred million, which permits fine-tuning on hardware with limited resources.

We load the Tiny Shakespeare dataset, comprising 100 lines of text. Each line of text is tokenized with truncation and padding to a sequence length of 256. The result is converted to a PyTorch dataset of input IDs and an attention mask. Labels are set equal to input IDs for causal language modeling.

The Trainer is configured with a batch size of 2, gradient accumulation over 4 steps (yielding an effective batch size of 8), a learning rate of \(2 \times 10^{-4}\), mixed precision (fp16), and 50 gradient steps. Logging occurs every 10 steps, and checkpointing occurs every 50 steps. A data collector places inputs and targets into tensors.

\subsection{Fine-Tuning the Attention Router}

The core of our approach is the AttentionRouter class. It takes hidden size (the size of token embeddings), number of experts, and a new default top k as 8. Internally, it projects each token into a lower qk-dim–dimensional query space by a learnable nn.Linear(hidden-size, qk-dim). Each expert is stored in self.expert embeddings as its own qk-dim-vector. These two components queries and expert centroids are the core of our routing attention. 

We initialize the weights and the query projection uses a Gaussian initialization (mean 0, std 0.02), and expert embeddings are either uniformly or normally distributed based on the init expert centroids flag. This makes sure that the keys and queries start in approximately the same scale before any training.

Routing probabilities are computed in compute router probabilities(). We reshape 2 dimensional inputs (batch×hidden) to 3 dimensional (batch×seq×hidden) when needed. We project tokens, and then we L2-normalize to avoid dot products drifting. We reshape queries and keys to the expected shape and use flash attn func, reshaping the output to (batch×seq×num experts). Otherwise, we compute the normal scaled dot-product, where T is the temperature. :

\begin{equation}
\mathrm{scores} \;=\; \frac{Q\,K^\top}{\sqrt{\mathrm{qk\_dim}}}
\;, \qquad
P \;=\; \mathrm{softmax}\!\Bigl(\frac{\mathrm{scores}}{T}\Bigr)
\end{equation}

The result is a proper probability distribution over experts for each token. The forward() function places inputs on the router device, invokes compute router probabilities() function, and then takes the log (with small epsilon) to match the seen "router scores" interface to the MoE block. Log-taking is useful numerically when the resulting scores would be combined with downstream expert outputs. 

To add our new router to Qwen, replace-qwen routers() walks through every sub-module looking for the old gate and an.experts list like we have done previously while implementing our Linear Router mechanism. It prints each location, looks at either module.gate.weight.shape or defaults to expert.gate projection to find (number of experts, hidden size), and reads the original top k or defaults to 8. Then it creates an Attention Router with those sizes and replaces both module.gate and, if present, module.shared expert gate. It keeps a count of the number of replacements made. 

Next, we once more wrapped the model within PEFT's LoRA but this time included the new query projection to the target modules. Thus, the adapter layers will not just learn low-rank updates over the typical attention projections but also our routing query projection, enabling us to parameter-efficiently fine-tune the router itself.

The \texttt{TrainingArguments} mirror our previous experiment: we use a per device batch size of 2 with gradient accumulation over 4 steps (for an effective batch size of 8), a learning rate of $2\times10^{-4}$, mixed-precision training (\texttt{fp16}), and a fixed budget of 50 total gradient steps. Metrics are logged every 10 steps.

\section{Results and Evaluation}

Each router was characterized through 1,024 inference runs, with 5 repetitions per router, and results were averaged.

\begin{table*}[htbp]
\caption{Comparison of MoE Router Configurations on Model Parameter Size.\protect\\
\footnotesize Parameter count reflects the total number of trainable parameters within each router module. Input Configuration: Batch Size = 16, Sequence Length = 128, Hidden Size = 768}
\label{table_router_param_size}
\begin{center}
\resizebox{0.3\textwidth}{!}{%
\begin{tabular}{|l|c|}
\hline
\textbf{Router} & \textbf{Parameter Size} \\
\hline
Linear         & 6,144   \\
Attention      & 49,664  \\
MLP            & 101,000 \\
Hybrid         & 55,810  \\
Hadamard       & 101,000 \\
Hash           & 0       \\
\hline
\end{tabular}%
}
\end{center}
\end{table*}

\subsubsection{Random Initialization}
\begin{table*}[htbp]
\caption{Comparison of MoE Router Configurations on Routing Behavior and Latency with Randomly Initialized Hidden states.\protect\\
\footnotesize Input Configuration: Batch Size = 16, Sequence Length = 128, Hidden Size = 768, Average of 1024 runs}
\label{table_router_latency_entropy} 
\begin{center}
\resizebox{0.8\textwidth}{!}{
\begin{tabular}{|l|c|c|c|}
\hline
\textbf{Router} & \textbf{Token Latency (ms)} & \textbf{Entropy} & \textbf{Mean Top-k Prob.} \\
\hline
Linear         & 0.07    & 1.9513 & 0.2129 \\
Attention      & 0.29    & 2.0793 & 0.1273 \\
MLP            & 0.23    & 2.0769 & 0.1361 \\
Hybrid         & 0.58    & 2.0476 & 0.1686 \\
MLP-Hadamard   & 0.88    & 1.1003 & 0.4152 \\
Hash           & 85.0    & 0.0000 & 0.5000 \\
\hline
\end{tabular}%
} 
\end{center}
\end{table*}

When we fine tuned linear and the attention on the Qwen model we saw that the loss was decreasing but had difficulty using this on datasets.

\begin{figure}[h]
    \centering
    \includegraphics[width=1.0\linewidth]{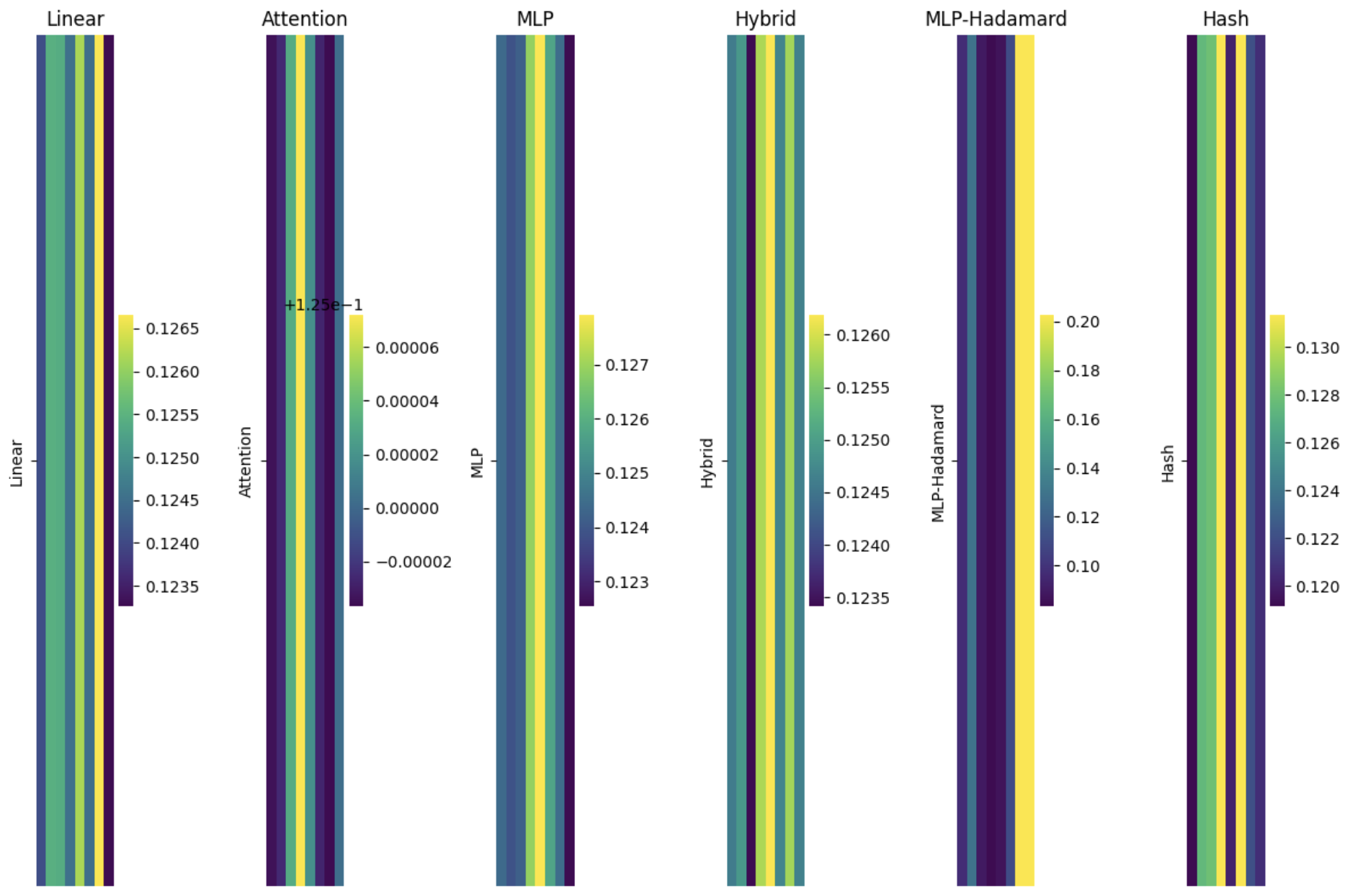}
    \caption{Visualization of Router Output Probabilities Across an Input Sequence. Each heatmap displays the probability distribution assigned by the respective router (Linear, Attention, MLP, Hybrid, MLP-Hadamard, and Hash) to eight experts (implicit x-axis) for a sequence of input token representations (y-axis). Color intensity corresponds to the probability, with scales provided by the color bars.}
    \label{fig:routing_prob_heatmap}
\end{figure}

Figure \ref{fig:routing_prob_heatmap} shows a comparison of how different router architectures distribute routing probabilities across the available experts for a sequence of input tokens. The Linear, Attention, MLP, and Hybrid routers have a relatively smooth probability assignments, showing a softer allocation of tokens. In contrast, the MLP-Hadamard and Hash router a concentrated probability distribution, often strongly favoring a smaller subset of experts, this might be its design combining MLP expressiveness with Hadamard-influenced feature interaction.
 
\begin{figure}[h]
    \centering
    \includegraphics[width=1.0\linewidth]{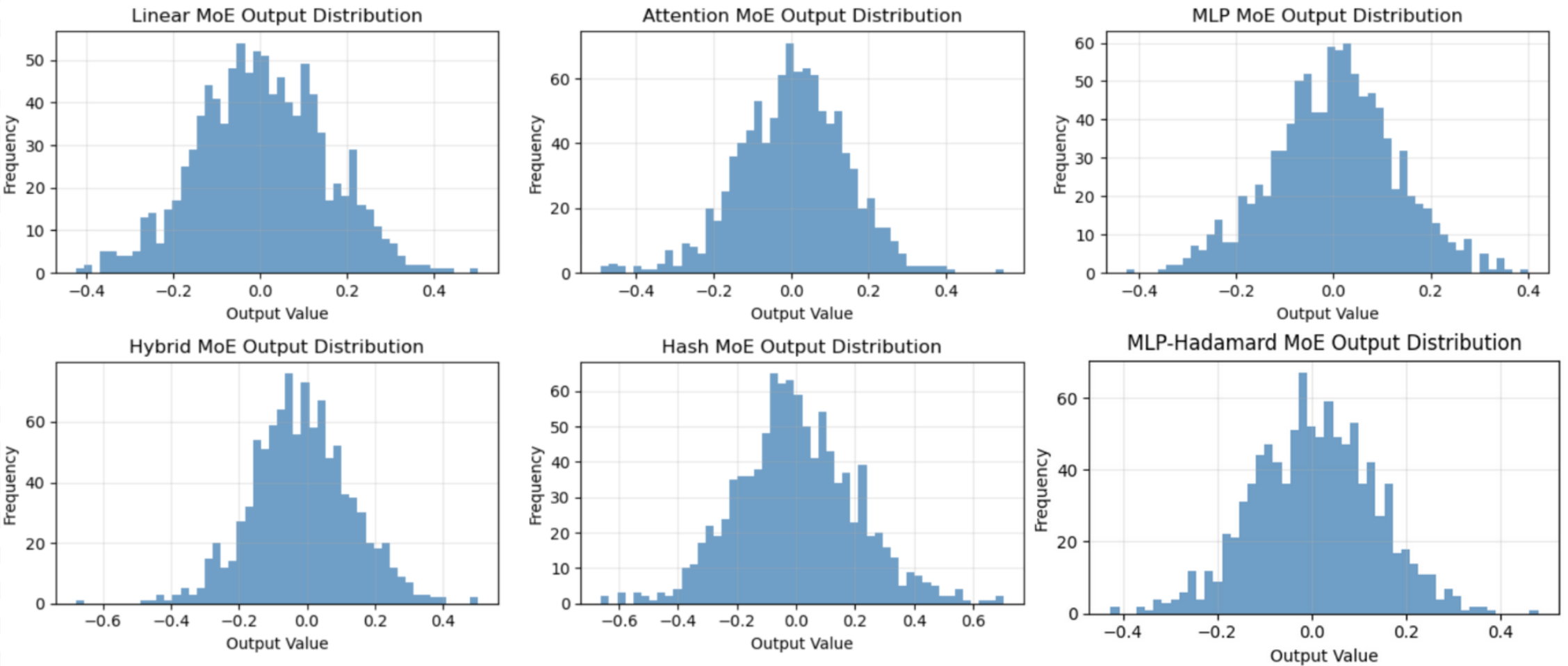}
    \caption{Histograms of MoE Layer Output Value Distributions for Different Router Configurations. Each histogram shows the frequency distribution of the aggregated output values from the MoE layer when using the specified router (Linear, Attention, MLP, Hybrid, MLP-Hadamard). These distributions show the impact of different routing strategies on the final activations passed to subsequent layers.}
    \label{fig:moe_output_distribution}
\end{figure}
The distributions of the MoE layer's output activations, in Figure \ref{fig:moe_output_distribution}, show how different routing mechanisms change the information passed forward. Most routers (Linear, Attention, Hybrid, MLP-Hadamard) produce output distributions that approximate a normal distribution, which is consistent with the combination of multiple expert outputs and add the Central Limit Theorem. Notably, the MLP router, results in an output distribution with a significantly different scale and a tighter concentration around zero compared to the others. This suggests that the MLP router, potentially due to its increased representational capacity or specific learned routing patterns during this un-fine-tuned BERT-based evaluation, might be leading to a more selective or compressive effect on the activations.
 
\begin{figure}[h]
    \centering
    \includegraphics[width=1.0\linewidth]{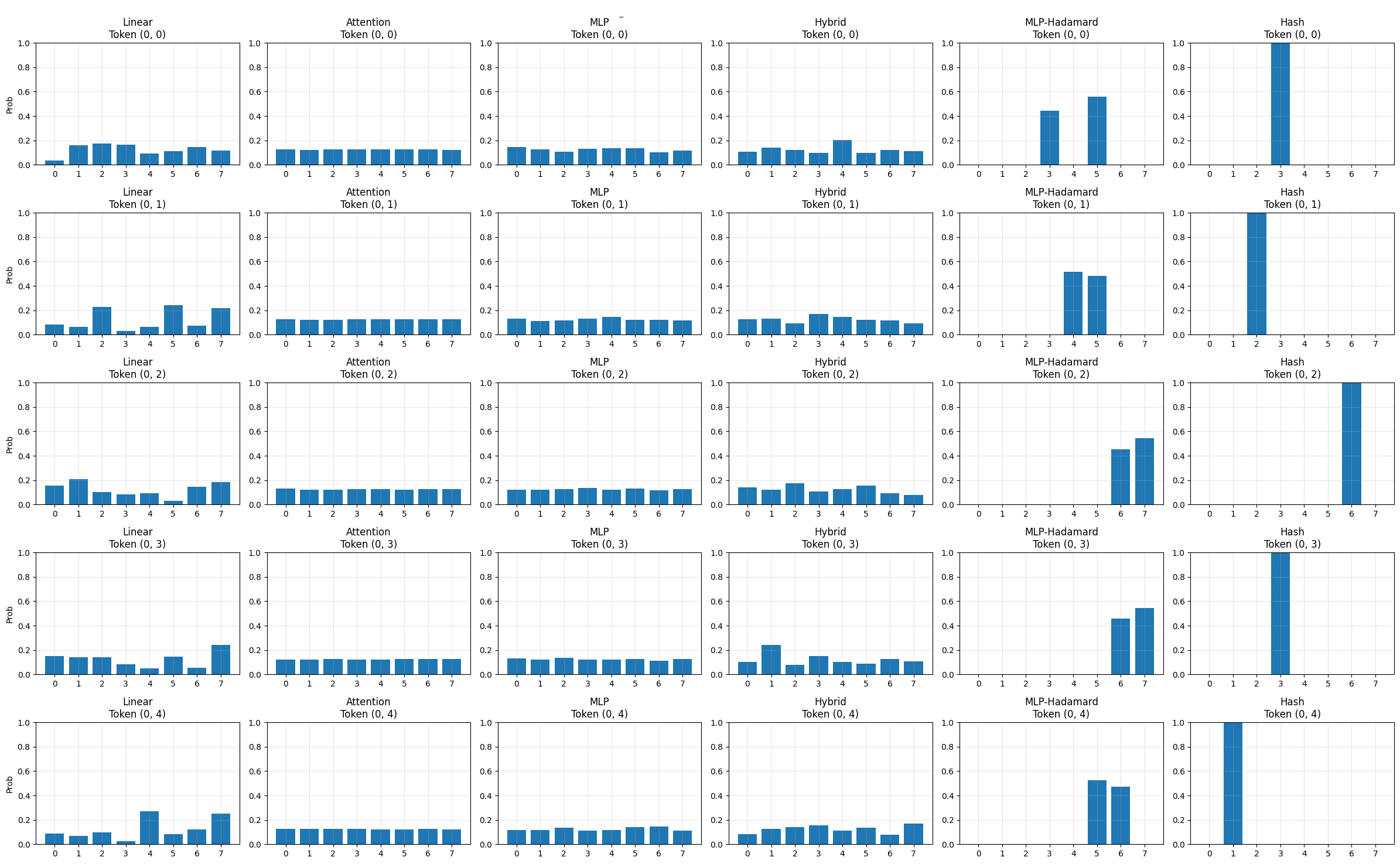}
    \caption{Per-Token Expert Routing Probability Distributions for Various Routers. This figure displays bar charts illustrating the probability assigned by each router (Linear, Attention, MLP, Hybrid, MLP-Hadamard) to each of the 8 experts (x-axis) for five different example tokens (Token (0,0) to (0,3)). The y-axis represents the probability. This shows the fine-grained routing decisions made by each architecture for specific inputs, Notice the hard k=2 expert routing characteristics of Hadamard.}
    \label{fig:per_token_routing_distribution}
\end{figure}
Figure \ref{fig:per_token_routing_distribution} shows the expert probability assignments for individual tokens across the different router architectures. The Linear, Attention, MLP, and Hybrid routers generally are more distributed set of probabilities across the experts for any given token, although preferences for certain experts can still be observed. This makes sense with their design allowing for softer, more expressive routing. The MLP-Hadamard router, however, has a distinctly different behavior, consistently assigning high probabilities to a very sparse set of experts (often two) for each token, indicating a more deterministic and concentrated routing strategy.

\section{Discussion}

The \texttt{LinearRouter} offered computation efficiency but had limited expressive power for complex token-to-expert mappings. The \texttt{MLP} router's capacity had varied routing dynamics, useful for tasks needing specific token handling. Deterministic routers, like \texttt{Hash} and the initial \texttt{HadamardRouter}, ensure predictable expert selection but can cause poor load balancing without design considerations. The \texttt{HadamardRouter}'s top-k constraint mitigated this, achieving controlled sparsity with diversity. Results confirm low routing entropy correlates with concentrated top-k probabilities, leading to deterministic behavior. Higher entropy promotes broad expert utilization, which balances load but can reduce routing specificity. Router evaluations with random inputs did not consistently show these performance characteristics. Using pretrained BERT weights for initialization showed more representative routing behaviors. This the need for appropriate initializations for router assessment. The \texttt{MLP} router's performance under such conditions indicates complex routers, despite higher computation needs, can achieve selective, efficient expert usage with correct initialization and constraints.

Router architecture evaluation revealed design trade-offs and performance. The \texttt{HadamardRouter} development shows this the most. We initially designed the \texttt{HadamardRouter} as parameter-free. It used a Hadamard matrix for computation simplicity and deterministic routing. Early tests showed a problem: the router directed all tokens to a single expert. This caused load imbalance and stopped MoE sparsity benefits.

We changed the Hadamard matrix. It functioned as a structural regularizer to impose orthogonality and prevent expert collapse, not as a direct router. Second, we implemented a \(k\)-experts hyperparameter to limit expert selection per token. These changes made a more stable and predictable module. The modified \texttt{HadamardRouter} routed tokens to two experts.

\subsection{Fine-Tuning the Linear Router}
Fine-tuning with LoRA adapts the pre-trained base model to the Shakespearean text domain without retraining the entire model. This also is a test for the router replacement. A monotonic loss curve shows the correct functioning of the new routing mechanism.

During the initial ten steps of fine-tuning, the training loss decreases from 1.9086. This initial loss value means that the model's next-token predictions do not yet reflect the patterns in the dataset.

By step twenty, the loss decreases to 0.1569. This is where adapter weights adapt to token sequences in the 100 lines. At step thirty, the loss reaches 0.0027. This loss value high model confidence in its predictions and suggests memorization of training instances.

From step forty to step fifty, the loss changes from 0.0009 to 0.0006. These changes mean there are now diminishing returns from further optimization, and the model converges on the training set. This means that the fine-tuning pipeline functions as designed.

The decrease in loss from approximately 1.9 to near zero within thirty updates on 100 lines of text make it look like the model is overfitting. Cross-entropy losses in the \(10^{-3}\) range mean that the model memorizes sequences in the training set rather than learning generalizable patterns.

\subsection{Fine-Tuning the Attention}

Early in training (Step 10, loss = 9.645), the model’s parameter including both the Transformer layers and the newly introduced attention-router projections are essentially at their random initial values. At this stage, the next-token prediction task is barely better than chance, which is exactly what you’d expect when neither the expert assignments nor the core language model weights have had time to adapt to the data.

By twenty steps (Step 20, loss = 6.0565), we begin to see the effects of learning kicking in both within the routing mechanism and also in the LoRA adapters. The router is beginning to specialize which expert it will consult for each token, and the base model parameters are shifting towards known Shakespearean patterns. That decrease of roughly 3.6 in loss over the span of ten steps represents good early progress. At the halfway mark (Step 30, loss = 3.5086), the training loss has halved once more. Here the model is refining its token-to-expert mappings and more confidently predicting the next character in each sequence. By step forty (Step 40, loss = 1.7095), most of the learning is primarily on this extremely tiny 100-line dataset. The model is now placing most of its probability mass on the correct next token, but with a tiny error buffer. This means that while it has grasped the broad statistical trends, a little bit of room remains for more calibration. Finally, at fifty updates (Step 50, loss = 0.7318), the model has strong predictive confidence its loss has decreased below one. Unlike the linear router experiment, which drove loss towards zero (making it look like it was just near-perfect memorization), our attention-based router follows a slower, more gentle decline. This is due to it learning more thoughtfully and retaining some uncertainty rather than overfitting the toy corpus instantly.

\subsection{Comparison with Prior Work}
\label{subsec:comparison_prior_art}
The \texttt{LinearRouter}, similar to designs by Shazeer et al. \cite{shazeer2017outrageously}, Fedus et al. \cite{fedus2022switch}, and Mixtral \cite{jiang2024mixtralexperts}, showed efficiency (0.07ms latency, Table I). Its entropy (1.9513) was less than the \texttt{AttentionRouter} (2.0793). This shows routing with less adaptation and motivates exploring other mechanisms.

The \texttt{AttentionRouter} and \texttt{MLPRouter} aim for expressiveness beyond linear-type projections. The \texttt{MLPRouter} with pretrained BERT embeddings (Figure \ref{fig:moe_output_distribution}) showed an output distribution with specific characteristics, using input features for expert selection. \texttt{AttentionRouter} fine-tuning on Qwen1.5-MoE showed a learning trajectory (loss 9.645 to 0.7318, 50 steps) different from \texttt{LinearRouter}'s overfitting. Attention's inductive biases may benefit routing and generalization, seeing similar to soft-mixture \cite{puigcerver2023poolingformer}.

The \texttt{MLP-Hadamard} router is a new design. Unlike hash-type routing \cite{roller2021hash}, it uses Hadamard interaction within an MLP, regularized and constrained to top-k selection (two experts). It got the minimum entropy (1.1003) and maximum mean top-k probability (0.4152) in static-condition tests. This adds a mechanism for sparsity and expert specialization with structure, differing from standard top-k selection.

\subsection{Limitations and Challenges}
One of the biggest challenges we had during the fine-tuning was adapting the Qwen1.5MoE model architecture to work successfully with our custom router implementations. The Qwen1.5MoE model uses a specific gating mechanism that needed to be completely replaced. To fix this, we built a helper function that ``walks'' the Qwen1.5MoE graph, spots every router, and swaps in our router implementation. First, we had to reverse-engineer Qwen1.5MoE's internal gating mechanism, identify exactly where and how each router connected into its specialists, and then modify our new router to slot in without disrupting any of the surrounding wiring. We created a \texttt{replace\_qwen\_routers()} utility that recursively searches every sub-module for the \texttt{giveaway.router} and \texttt{.experts} properties, saves the dimensions and routing hyperparameters of the original router, and substitutes it with our AttentionRouter in place of the old gate. By simply replacing the router objects without touching expert layers, remaining connections, and forward-pass logic even in cases where multiple gates share the same experts , we are able to preserve the integrity of the original design and simply insert our attention-based routing mechanism.

Additionally, dealing with the GPTQ quantization format introduced a whole new level of complexity, since our AttentionRouter just couldn't inherit the compressed 4-bit weight tensors in their original form. When we tried to load a fine-tuned GPTQ checkpoint, PyTorch would give ``unexpected keys'' (e.g., \texttt{query\_proj.lora\_A.default.weight}) error precisely because our custom router parameters didn't exist within the original MoE structure.

Besides that, linear layers quantized reveal non-standard-shaped and bit-packed weights, and naively copying dimensions would have broken downstream matrix multiplies. To fix this, we added a dimension-sniffing routine that introspects floating-point and GPTQ quantized layers both pulling in/out features regardless of whether the data is in \texttt{weight.data} or a compressed buffer , and we changed our AttentionRouter constructor to accept those dimensions explicitly. That way, whether the parent model is saved in a GPTQ format or not, our code correctly allocates and initializes each projection matrix, aggregates in LoRA adapters when present, and avoids missing-key or shape-mismatch errors during the forward pass.

Scaling our own AttentionRouter on top of the already enormous Qwen1.5MoE base foundation produced its own memory and performance problems. With \(2.7\) billion base parameters and LoRA adapters to boot, GPU memory was not enough, and the extra tensors used to calculate the attention scores just added to the pressure. Benchmarks that handled hundreds or thousands of examples in parallel would sometimes overwhelm even high-end cards like A100, and inference latency increased dramatically especially on devices where FlashAttention kernels were not available. In order to keep experimentation moving, we added ``quick test'' modes that subsample data. These changes allow us to guarantee correctness without incurring the full memory or speed cost of a production run each time.

\section{Conclusion}

We created a framework that did a comparative analysis of Linear, Attention, MLP, Hybrid, MLP-Hadamard, and Hash routers, characterizing their performance using latency, parameter count, routing entropy, and expert utilization. Our experiments, conducted with both randomly initialized and pretrained BERT embeddings, and through fine-tuning on the Qwen1.5-MoE model, showing the operational characteristics and trade-offs inherent in each design.

We see that simpler routers like the LinearRouter achieve low latency, while more complex architectures such as the MLP and Attention routers offer increased representational power, leading to more nuanced routing decisions. The new MLP-Hadamard router provided a structured approach to achieving sparse expert activation. A key part of this report was the successful integration and fine-tuning of our custom routers, including the AttentionRouter, within the quantized Qwen1.5-MoE architecture. This process involved large challenges from model introspection, parameter adaptation for quantized layers, and managing computational resources. Our report shows the role of router design in MoE performance and the importance of context-aware evaluation using pretrained models.

\section{Future Work}

We want to next look at more adaptive routing mechanisms. These are routers that dynamically adjust their top-k selection strategy or internal parameters based on input token characteristics, layer depth, or even real-time feedback on expert load and performance. We could use reinforcement learning to train routers for optimal long-term task performance and resource utilization.

We also aim to have more evaluations of the developed routers on a larger range of downstream tasks and larger, more diverse datasets. Further exploration into the interplay between router design and expert architecture specialization could also yield significant performance gains.

We will optimize the overhead of more complex routers like the Attention and MLP variants. We could use distillation from a complex router to a simpler one, or developing more efficient implementations could make these expressive routers more practical.

We also are interested to further analyze the more theoretical parts of different routing strategies, especially the dynamics of expert specialization and the conditions that lead to optimal load balancing. We think that having a deeper mathematical understanding of how routers split up the feature space could help us the design of next-generation MoE systems.

\section{Contributions}
\begin{table}[H] 
\centering
\caption{Author Contributions. All authors contributed equally (33.3\% each) to the overall research and writing.}
\label{tab:contributions}
\begin{tabularx}{\columnwidth}{@{} l c >{\RaggedRight\arraybackslash}X @{}}
\toprule
\textbf{Author} & \textbf{Share} & \textbf{Specific Contributions} \\
\midrule
Dan Harvey (dyh2111) & $\sim$33.3\% & Worked on Report, Generated BERT results, Quantized Models, Developed Hadamard Router, Worked on Qwen \\
\addlinespace 
George Weale (gmw2143) & $\sim$33.3\% & Worked on Report, Started the Initial Framework and the initial routers, and Mistral dissection. \\
\addlinespace
Berk Yilmaz (by2385) & $\sim$33.3\% & Worked on Report, Implemented more advanced routers, Fine tuned models, and worked on Qwen \\
\bottomrule
\end{tabularx}
\end{table}

\section{Thank you!}
We really really enjoyed this project, this was the hardest deep learning project we have worked on. This project made us dive deeper into deep learning than all of us had previously done. Getting to the core of these open source models and tearing them apart and building them up again gave us so much appreciation as working with these fully torn down models was very difficult. All three of us really appreciated the flipped class room approach and how we were able to learn from other students presenting and how were we able to present to the class. Thank you Professor Kostic, thank you William Ho, and thank you Ian Li for this quarter!

\section{Supplemental}

We tried many routers, the six in the main report we found to be the most useful, but one that just didnt make the cut as it was very hard to work with.
\begin{table}[H] 
\centering
\caption{Supplemental Router Performance Metrics with Pretrained BERT Initialization}
\label{tab:supplemental_router_metrics}
\resizebox{\columnwidth}{!}{%
\begin{tabular}{@{}lcccc@{}}
\toprule
\textbf{Router Type} & \textbf{Entropy} & \textbf{Mean Top-k Prob.} & \textbf{Output STD} & \textbf{Auxiliary Loss} \\
\midrule
Linear           & 2.0680 & 0.1486 & 0.03083 & 0.00126 \\
Attention        & 2.0793 & 0.1271 & 0.02950 & 0.00125 \\
MLP              & 2.0793 & 0.1277 & 0.06820 & 0.00125 \\
Hybrid           & 2.0767 & 0.1368 & 0.03197 & 0.00126 \\
Hash             & 0.0000 & 0.5000 & 0.03213 & 0.00229 \\
Self-Supervised  & 0.6146 & 0.5000 & 0.02987 & 0.00218 \\
\bottomrule
\end{tabular}%
}
\end{table}
\subsubsection{Self-Supervising Router}
We also wanted to look at a self-supervised router to use input representations for making routing decisions. The main idea is that features learned from self-supervised pretext tasks can provide more information for determining expert assignments. In our implementation, the self-supervised router first processes the input token representation $x \in \mathbb{R}^d$ using a feature extractor network, $f_{\theta_{ss}}(x)$. The parameters $\theta_{ss}$ of this network can either be pre-trained using a self-supervised objective $\mathcal{L}_{ss}$, or they can be trained concurrently with the main model.

\begin{equation}
x'_{ss} = f_{\theta_{ss}}(x)
\end{equation}

The  features $x'_{ss} \in \mathbb{R}^{d'_{ss}}$ extracted by this network are then passed to a subsequent routing head to compute the final expert probabilities:

\begin{equation}
p(e|x) = \text{softmax}(W_{route} \cdot x'_{ss} + b_{route})
\end{equation}

where $W_{route} \in \mathbb{R}^{n \times d'_{ss}}$ and $b_{route} \in \mathbb{R}^{n}$. We see that by utilizing representations refined through self-supervision, this router could see more more structural properties of the tokens, having a more effective partitioning of the problem space. As it $\mathcal{L}_{ss}$ takes the form:

\begin{equation}
\mathcal{L}_{ss} = -\log \frac{\exp(\text{sim}(x'_{ss}, x'^{+}_{ss})/\tau_{ss})}{\sum_{j} \exp(\text{sim}(x'_{ss}, x'^{-}_{ss,j})/\tau_{ss})}
\end{equation}

where $x'^{+}_{ss}$ is a positive pair to $x'_{ss}$, $x'^{-}_{ss,j}$ are negative pairs, $\text{sim}(\cdot)$ is a similarity function (e.g., cosine similarity), and $\tau_{ss}$ is a temperature parameter for the self-supervised task.

\end{document}